\documentclass[11pt]{article}

\usepackage[preprint]{acl}

\usepackage{times}
\usepackage{latexsym}
\usepackage{amsmath}
\usepackage{amssymb}
\usepackage{enumitem}
\usepackage{hyperref}
\usepackage{mathtools}
\usepackage{booktabs}
\usepackage{csquotes}
\usepackage{breqn}
\usepackage{amsmath}
\usepackage[capitalize]{cleveref}

\usepackage{tikz}
\usetikzlibrary{positioning, arrows.meta, fit, backgrounds, calc, shapes.geometric}
\usepackage{pgfplots}
\usepgfplotslibrary{groupplots}
\pgfplotsset{compat=1.18}

\usepackage[T1]{fontenc}

\usepackage[utf8]{inputenc}

\usepackage{microtype}

\usepackage{inconsolata}

\usepackage{graphicx}

\usepackage{xspace}

\newcommand{\MMTM}{\texttt{MMTM}\xspace}
\newcommand{\Witness}{\texttt{MMTM}\xspace}

\newcommand{\Anon}{\textsc{\textbf{anonymized for review}}\xspace}

\title{\Witness: Tri-Modal Topic Modeling for Long-Form Video via Similarity-Gated Fusion}

\author{Ali Abusaleh, Bhuvanesh Verma, Prof. Dr. Alexander Mehler \\ 
Text Technology Lab (TTLab),  \\
         Goethe University Frankfurt\\
         \{a.abusaleh,verma,mehler\}@em.uni-frankfurt.de
         }

\newcommand{\AM}[1]{\textcolor{black}{#1}}

\begin{document}
\maketitle

\begin{abstract}
We introduce \Witness, a modular pipeline for topic discovery in long-form video that integrates speech recognition, audio and visual embeddings, and BERTopic clustering through a deterministic similarity-gated fusion. Evaluated cross-lingually on German (\textit{Tagesschau}) and English (NBC) broadcast news, \textbf{joint tri-modal modeling} substantially improves topic quality: noise drops from 0.27 to 0.06, transition rate from 0.70 to 0.21, and normalized entropy rises from 0.84 to 0.92, indicating more coherent and temporally stable topics. Cluster validity (Calinski--Harabasz) improves by 5--12$\times$ across embedding spaces. Lexical coherence (NPMI) rises from 0.77 to 0.86 on German but is corpus-dependent and does not transfer to the shorter NBC broadcasts. We release the pipeline code and a human-validated 54-hour multimodal video topic corpus with dual-annotator visual evaluation and LLM-assisted labeling.\footnotemark
\end{abstract}
\footnotetext{Code and data release upon acceptance.}

\section{Introduction}

Long-form video, including broadcast news and live streams, constitutes a growing share of online information \cite{Budzinski2021-sl}.
Understanding their topical structure is critical for media analysis, disinformation research, and content retrieval \cite{park:et:li:2007, fu:et:al:2025:misinformation}.
However, most topic models rely on text alone \citep{zheng2014topic, lokmanoglu2025vistopics}, or at most two modalities \citep{gonzalez2024neural}, ignoring acoustic and visual cues humans use to parse stories and segment discourse.
%
Visual cues such as composition, color, and symbolic action \citep{zhu2013videotopic,lokmanoglu2025vistopics} and acoustic signatures, such as ambient soundscapes, emotional valence or tonal shifts  \citep{kim2009acoustic,hu2014latent}, function as subtextual framing devices that clarify intent and urgency in ways that mere transcription cannot reflect.
Consequently, there is a compelling need to integrate these modalities to capture the 
semantic context inherent in the interplay between textual, acoustic, and visual information.

\begin{figure*}[h]
\centering
\includegraphics[width=0.9\textwidth, keepaspectratio]{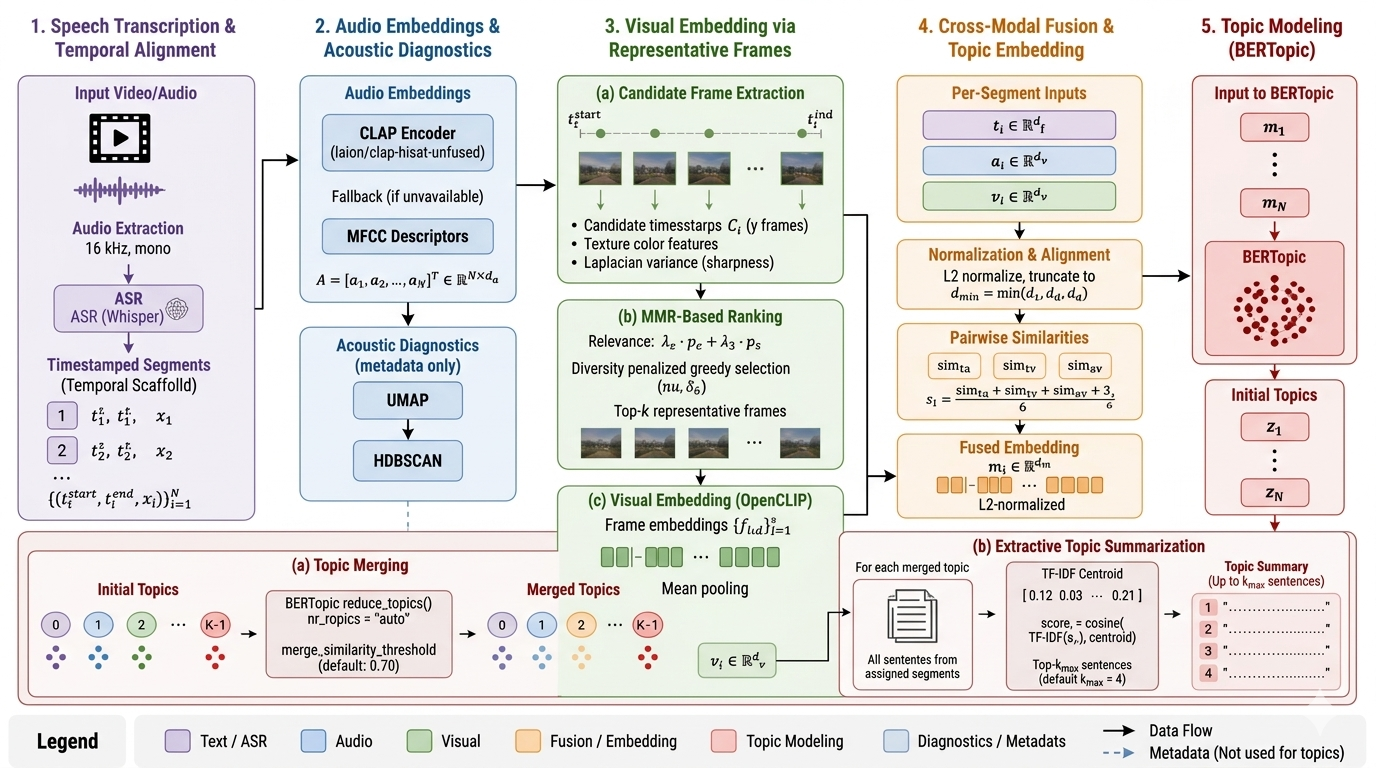}
\caption{\AM{Architecture of \Witness. Visualization created using \href{https://paper-banana.org/}{PaperBanana} based on the authors' design.}}
\label{fig:MMTM}
\end{figure*}

%
We introduce \AM{\textbf{\Witness (Multimodal Topic Modeling in Long-Form Video via Similarity-Gated Fusion)}}, a modular pipeline that bridges this gap:
%
It extracts segment-level text, audio, and visual representations with pretrained encoders, fuses them via a deterministic similarity gate, and clusters the joint embeddings into interpretable topics using BERTopic.
%
\AM{\Witness is modular:} 
\textcolor{black}{each processing stage, including speech transcription (Whisper), audio encoding (CLAP, Pyannote, MFCC), visual encoding (OpenCLIP, SigLIP, an OpenWebUI-based VLLM), and topic clustering (BERTopic), is independently replaceable without affecting the rest of the pipeline.
%
\AM{It supports configurable multimodal fusion weights} and \textbf{both unsupervised and seed-based weakly supervised topic modeling}.
%
%
%
%
\Witness uses an unsupervised, inference-only setting without predefined labels, training splits, or fine-tuning. 
%
%
This minimizes risks of data leakage, overfitting, or train/test contamination, since every video is processed independently under identical conditions.}

A challenge for reproducible multimodal topic modeling is the lack of datasets with validated topic labels and aligned multimodal representations.
To address this gap, \textbf{we release \MMTM together with a multimodal video topic dataset} of roughly 54 hours of broadcast news, validated by a dual-annotator human study.
The annotation toolkit, including LLM-assisted topic labeling and visual intrusion/matching tests, will be released with this dataset.

We evaluate \Witness on \textbf{cross-lingual} German and English news corpora, comparing text-only and multimodal topic models.
Our results show that joint tri-modal modeling improves topic quality: noise drops from 0.27 to 0.06, transition rate from 0.70 to 0.21, and topic entropy rises from 0.84 to 0.92.
Cluster validity indices (\textit{Calinski-Harabasz}, \textit{Silhouette}, \textit{Davies-Bouldin}) improve by factors of 4--12, indicating stronger cross-modal structure.
Lexical coherence (NPMI) rises from 0.77 to 0.86, indicating improved semantic interpretability through multimodal evidence.
%
Our main contributions are threefold:
\begin{enumerate}
    \item An open-source, modular pipeline for multimodal video topic modeling that combines ASR, CLAP, and OpenCLIP embeddings through similarity-gated fusion, followed by BERTopic clustering and 
    weak supervision;
    \item A multimodal video topic dataset with hu\-man-va\-li\-da\-ted annotations, including dual-annotator visual evaluation and LLM-assisted topic labeling, 
    {available here}\footnote{upon manuscript acceptance};
    \item An evaluation demonstrating improved structural and lexical topic quality through multimodal fusion.
\end{enumerate}

\section{Related work}

Topic modeling traditionally extracts latent thematic structure from text, using methods ranging from probabilistic models \cite{blei2003latent} to neural approaches.
However, text-only analysis is inherently limited, as real-world information is multimodal.
Latent topic interpretation can be enriched through multimodal context.
Consequently, attention has shifted toward multimodal topic extraction.
\citet{blei2003modeling} pioneered this shift with Corr-LDA, jointly modeling images and captions.
Other early probabilistic approaches include tr-mmLDA \citep{putthividhy2010topic}, which learns modality-specific topic sets and predicts one from the other via regression.
%
Similarly, \citet{jia2011learning} model visual and textual words as multinomial distributions within a multimodal document random field to learn shared cross-modal topics.
%
\citet{zheng2014topic} proposed an early neural autoregressive topic model for multimodal data, jointly modeling an image's visual words, annotation words, and class labels.
More recently, NN-based methods for topic modeling have emerged.
%
Most rely on neural variational inference \citep{miao2016neural}, typically via variational autoencoders (VAEs), as in ProdLDA \citep{srivastava2017autoencoding} and the Embedding-based Topic Model (ETM) \citep{qiang2017topic}.
%
These neural approaches remain largely restricted to text.
%
Addressing this gap, BERTopic \citep{grootendorst2022bertopic} introduced a shift by decoupling document embedding from topic representation.
%
This paved the way for multimodal extensions by enabling independent multi-source representations to be integrated before clustering.
Following the success of BERTopic, many studies adopted embedding-based topic modeling for multimodal settings.
%
For instance, \citet{prakash2023promptmtopic} introduced PromptMTopic, an unsupervised multimodal framework that uses LLMs to jointly analyze memes' visual and textual content.
%
\citet{gonzalez2024neural} added two algorithms: 
%
Multimodal-ZeroShotTM combines image embeddings with bag-of-words reconstructions, while Multimodal Contrast uses contrastive learning to align modalities in a shared topic space.
%
They also introduced two evaluation metrics, image-embedding coherence (IEC) and image-embedding pairwise similarity (IEPS), which we adopt in our evaluation.
%
Extending topic modeling to video data, \citet{lokmanoglu2025vistopics} introduced VisTopics. 
%
It extracts keyframes from news videos, generates captions with LLMs, and applies LDA to the resulting text.
%
While motivating our focus on news video, their model remains text-based; we instead use modality-specific encodings to learn latent features across modalities and derive a unified multimodal topic distribution.

\AM{Tri-modal topic modeling over text, audio, and visual streams remains largely unexplored.}
%
Prior multimodal topic models either use image-caption pairs \citep{blei2003modeling, gonzalez2024neural} or reduce video to text via captioning before applying LDA \citep{lokmanoglu2025vistopics}.
%
We therefore compare against the strongest available baseline: text-only BERTopic with the same transcripts, encoder, and clustering settings, isolating the effect of multimodal fusion.


\section{Methodology \& Dataset}

\AM{In this section, we describe the video corpus constructed for topic modeling and the \Witness architecture used to process it (see \autoref{fig:MMTM}).}

\subsection{Transcription and Temporal Alignment}
We extract audio at 16 kHz mono for compatibility with downstream ASR models.
We utilize Whisper \cite{radford:2022:whisper} for transcription, generating timestamped text segments, \AM{where $x_i$ denotes the transcript text of the $i$-th segment and $N$ is the total number of segments:}:
\begin{equation}
\{(t_i^{\text{start}}, t_i^{\text{end}}, x_i)\}_{i=1}^{N}
\end{equation}
ASR segmentation serves as the primary temporal scaffold, as lexical boundaries align more accurately with semantic transitions than fixed-duration windows.
To mitigate the downstream propagation of substitution and deletion errors, we incorporate frequency-based lexical filtering.
%

\subsection{Audio Embeddings} 

For each ASR-aligned segment we extract an audio embedding using a CLAP-based encoder \cite{elizalde:2022:clap} (\texttt{laion/clap-htsat-unfused}): 
\begin{equation}
A = [\textbf{a}_1, \textbf{a}_2, \dots, \textbf{a}_N]^{\top} \in \mathbb{R}^{N \times d_a},
\end{equation}
\AM{where $\mathbf{a}_i \in \mathbb{R}^{d_a}$ denotes the audio embedding of segment $i$, encoding acoustic information that is complementary to the corresponding ASR transcript.}
If the primary CLAP backend is unavailable, the pipeline automatically falls back to MFCC-based descriptors, ensuring uninterrupted coverage.

The audio, visual, and text representations are fused via a deterministic weighted concatenation with similarity-gated interaction terms (see \autoref{sec:fusion}) to produce the multimodal embedding $m_i$.
BERTopic then directly consumes \AM{the segment-level representations $\{m_i\}_{i=1}^{N}$} for topic assignment.
\textbf{Acoustic diagnostics and metadata.} Independently of topic modeling, we cluster the audio embeddings with UMAP \cite{mcinnes:et:al:2020:umap} and HDBSCAN \cite{Malzer:2020:hdbscan} to obtain speaker-style labels $\ell_i^{(a)}$, \AM{where $a$ ...}
These labels are used solely as metadata in the interactive timeline and for diagnostic checks; they do \emph{not} participate in topic discovery.

\subsection{Visual Embedding} 
To avoid transition artifacts (e.g., scene cuts and motion blur) common at segment boundaries while ensuring visual diversity and non-redundancy, we employ a two-stage frame selection strategy: candidate sampling followed by diversity-aware ranking.
\subsection*{Candidate Frame Extraction}
For each segment $s_i = (t_i^{\text{start}}, t_i^{\text{end}})$, we first generate a larger candidate pool of timestamps:
\begin{equation}
\begin{multlined}
C_i = \{ \tau_{ij} = t_i^{\text{start}} + \frac{j}{\gamma+1}(t_i^{\text{end}} - t_i^{\text{start}}) \mid j=1..\gamma \}
\end{multlined}
\end{equation}
The number of candidate timestamps $\gamma$ is set to $\gamma = \max(k, \alpha k, m_c)$, where:
    $k=5$ is the number of representative frames we aim to select per segment;
    $\alpha = 4$ is a multiplier that expands the candidate pool beyond $k$ (to $20$) so that the diversity‑aware ranking has enough material to choose from;
    $m_c = 8$ is a minimum candidate count, ensuring that even very short or static segments are sampled with at least eight frames.

This guarantees a sufficiently diverse candidate set before the ranking step.
For each candidate timestamp, we extract the corresponding frame and compute: (1) a texture-color feature descriptor based on resized grayscale patches and a normalized color histogram; and (2) a per-frame sharpness score using Laplacian variance as a proxy for visual quality, inversely related to blur.
\subsubsection{Representative Frame Ranking} 
%
To select the final $k$ representative frames, we use a greedy diversity-aware ranking inspired by maximal marginal relevance (MMR) \AM{\cite{Carbonell:Goldstein:1998}}.
%
Each candidate frame is scored by:
\begin{equation}
\operatorname{relevance}_i(j) = \lambda_c \rho_c(\tau_{ij}) + \lambda_s \rho_s(j) 
\end{equation}
where:
$\rho_c(\tau_{i,j}) = 1 - \min(1, |\tau_{i,j} - m_i| / (0.5 \cdot \Delta t_i))$ is center-of-segment preference (frames near segment midpoint $m_i$ score higher);
$\rho_s(j) \in [0, 1]$ is normalized Laplacian-based sharpness;
$\lambda_c, \lambda_s$ are fixed weights (typically $\lambda_c=0.7, \lambda_s=0.3$).
%
The ranking process greedily selects frames that maximize a diversity-penalized score:
\begin{dmath}
\text{score}(j \mid \mathcal{S}) = (1 - \nu) \operatorname{relevance}_i(j) - \nu \max_{j' \in \mathcal{S}} \operatorname{sim}(f_j, f_{j'})
\end{dmath}
where $\mathcal{S}$ is the set of already selected frames, $\operatorname{sim}$ denotes cosine \AM{similarity} in feature space, $\nu \in [0,1]$ controls the trade-off between diversity and representativeness, and previously selected frames are marked ineligible if similarity exceeds a deduplication threshold $\delta_d$ (typically 0.96).

\subsubsection{Visual Embedding Post-Selection}
%
Once the final $k$ representative frames have been selected and persisted, they are encoded using the OpenCLIP model \AM{(Base model: \texttt{ViT-B-32}, pretrained weights: \texttt{laion2b_s34b_b79k})} \cite{ilharco_gabriel_2021_5143773, cherti2023reproducible, Radford2021LearningTV, schuhmann2022laionb}. 
Assuming that segment $i$ contains frame embeddings $\{\AM{\mathbf{f}}_{ij}\}_{j=1}^{k}$, the segment-level visual descriptor is obtained by mean pooling, which ensures permutation invariance and reduces the risk of overfitting (\AM{$d_v$ is the dimensionality of the visual embeddings}):
\begin{equation}
{\mathbf{v}_i = \frac{1}{k} \sum_{j=1}^{k} \mathbf{f}_{ij}, \quad \mathbf{V} = [\mathbf{v}_1, \dots, \mathbf{v}_N]^{\top} \in \mathbb{R}^{N \times d_v}}
\end{equation}

\subsection{Cross-Modal Fusion and Embedding}
\label{sec:fusion}

After obtaining segment-level representations for the text, audio, and visual modalities, we combine them using a deterministic, similarity-gated fusion scheme.
Let \AM{$\mathbf{t}_i \in \mathbb{R}^{d_t}$, $\mathbf{a}_i \in \mathbb{R}^{d_a}$, and $\mathbf{v}_i \in \mathbb{R}^{d_v}$} denote the text, audio, and visual embeddings for segment $i$.
We first apply L2 normalization to all three modalities and then truncate them to dimensionality $d_{\min} = \min(d_t, d_a, d_v)$, $\mathbf{x}_i'\in \{\mathbf{a}_i', \mathbf{t}_i', \mathbf{v}_i'\}$:
\begin{equation}
\mathbf{x}_i' = \frac{\mathbf{x}_i^{[:d_{\min}]}}{\|\mathbf{x}_i^{[:d_{\min}]}\|_2} 
\end{equation}
Then, for $\mathbf{x}_i', \mathbf{y}_i'\in \{\mathbf{t}_i', \mathbf{a}_i', \mathbf{v}_i'\}$, $\mathbf{x}_i'\not=\mathbf{y}_i'$, we compute the three pairwise dot-product similarities:
\begin{equation}
\operatorname{sim}_{xy} = \mathbf{x}_i' \cdot \mathbf{y}_i' \in [0, 1]
\end{equation}
A scale factor
\begin{equation}
s_i = \frac{\operatorname{sim}_{ta} + \operatorname{sim}_{tv} + \operatorname{sim}_{av} + 3}{6} \in [0, 1]
\end{equation}
\textcolor{black}{\AM{rescales} the sum of the three pairwise similarities 
so that segments whose modalities are mutually consistent receive higher weight.}
%
Finally, we obtain a fused representation by concatenating the weighted modalities and their pairwise interactions:
\begin{dmath}
\mathbf{m}_i = \text{L2Norm}\Big[ w_t s_i \mathbf{t}_i' ; w_a s_i \mathbf{a}_i' ; w_v s_i \mathbf{v}_i' ; \mathbf{t}_i' \odot \mathbf{a}_i' ; \mathbf{t}_i' \odot \mathbf{v}_i' ; \mathbf{a}_i' \odot \mathbf{v}_i' ; \mathbf{t}_i' \odot \mathbf{a}_i' \odot \mathbf{v}_i'\Big],
\end{dmath}
%
$w_t, w_a, w_v \in (0,1]$ are hyperparameters (default: 0.34, 0.33, 0.33), $;$ denotes concatenation, and $\odot$ element-wise multiplication \AM{(Hadamard product)}.
%
The final embedding $\mathbf{m}_i$ is L2-normalized and used by BERTopic for topic assignment.

\paragraph{Acoustic diagnostics.} 
Independent of topic modeling, we cluster the segment-level audio embeddings $\{\mathbf{a}_i\}$ using UMAP for nonlinear dimensionality reduction, followed by HDBSCAN for density-based cluster discovery.
The resulting cluster labels serve as speaker-like metadata associated with each segment and are stored as $\text{seg}[\text{"speaker"}]$.
These labels are used exclusively for visualization and segment-level diagnostics; they do \emph{not} influence topic assignment or multimodal fusion.

\subsection{Topic Refinement} 

After using BERTopic to assign initial topic labels to each segment, we apply two optional refinement steps to improve topic quality and interpretability.
\paragraph{Automatic Topic Merging}
High-dimensional clustering often produces near-duplicate topics that differ only in a small number of top words.
We use BERTopic's built-in \texttt{reduce\_topics()} method with \texttt{nr\_topics="auto"} to merge topics based on the similarity of their word representations.
This reduces the total number of topics while preserving meaningful distinctions.
The merging procedure uses an internal similarity threshold (configurable; default: 0.70 via \texttt{merge\_similarity\_threshold} in the configuration).

\paragraph{Extractive Topic Summarization}

For each topic, we select up to $k_{\max}$ 
sentences (default: 4) from all segments assigned to that topic.
We use a TF-IDF centroid method,
$\text{score}_j = \operatorname{cosine}(\mathbf{s}_j, \mathbf{c})$,
where $\mathbf{c}$ is the average TF-IDF vector of all sentences assigned to the topic, and $\mathbf{s}_j$ is the TF-IDF vector of candidate sentence $s_j$.
We then select the top-$k_{\text{max}}$ sentences ranked by these scores.
%
If stop-word filtering removes all terms, as occurs in short or noisy \AM{samples}, we fall back to a token-frequency heuristic, scoring each sentence by summed term frequencies in the topic corpus (cf.\ \autoref{sec:appendix_config}).

\subsection{Dataset}
\label{sec:dataset}
To evaluate the effectiveness of \Witness, we use two longitudinal multimodal news corpora that vary in language, broadcaster, and editorial format.

\paragraph{Tagesschau (German).}
We curate a corpus of long-form \textit{Tagesschau} broadcasts, the flagship news program produced by ARD-aktuell in Hamburg, Germany.
We systematically collected daily episodes of \textit{Tagesschau} 20 Uhr from January 1, 2026, to April 21, 2026\footnote{\url{https://www.tagesschau.de/archiv/sendungen?filter=tagesschau_20_uhr}}, yielding approximately 54 hours of broadcast news. 
This collection includes one video per day, with occasional additional broadcasts for special coverage (e.g., New Year's Eve or Easter). This corpus serves as the primary evaluation dataset for the main experiments (\autoref{sec:results}) and ablation studies 
(\autoref{sec:ablation}).

\paragraph{NBC News (English).}
To assess cross-lingual generalization, we additionally use an English-language NBC News corpus \cite{lokmanoglu2025vistopics}, comprising 223\footnote{22 video from the original dataset has formatting issue} videos, or approximately 20 hours of material, collected between November 10, 2023, and October 13, 2024.
This corpus differs from \textit{Tagesschau} not only in language but also in broadcaster, editorial format, and temporal coverage, thereby providing a strong test of pipeline robustness beyond a single source. Results on this corpus are reported in \autoref{sec:english_results}.

\section{Results}
\label{sec:results}
 
Unless noted otherwise, results are reported on both corpora. All experiments run \Witness in a seed-based weakly supervised setting, with BERTopic guided by the 15 seed-topic groups in \autoref{sec:appendix_config}.

\subsection{Multimodal Fusion on Topic Structure}

\AM{By enabling the shift from text-only to multimodal topic modeling, \Witness substantially alters the statistical profile of the discovered topics as shown on ~\autoref{tab:structure}.}
On German, noise drops from 0.27 to 0.06, transition rate from 0.70 to 0.21, entropy rises from 0.84 to 0.92, and the number of topics grows from 19.1 to 25.9, indicating finer thematic distinctions than text alone can resolve. On English, the same pattern holds qualitatively: noise drops by 43\% and transition rate roughly halves.

\begin{table}[t]
\centering
\footnotesize
\setlength{\tabcolsep}{3pt}
\begin{tabular}{l rrrr}
\toprule
\textbf{Metric} & \textbf{Text} & \textbf{T+A} & \textbf{T+V} & \textbf{\Witness} \\
\midrule
\multicolumn{5}{l}{\textit{German}} \\
Noise $\downarrow$           & 0.273 & 0.076 & 0.087 & \textbf{0.057} \\
Transition $\downarrow$      & 0.699 & 0.390 & 0.278 & \textbf{0.207} \\
Entropy $\uparrow$           & 0.837 & 0.805 & 0.913 & \textbf{0.923} \\
Gini $\downarrow$            & 0.384 & 0.421 & 0.366 & \textbf{0.359} \\
\# Topics $\uparrow$         & 19.1 & 10.1 & 25.5 & \textbf{25.9} \\
\midrule
\multicolumn{5}{l}{\textit{English (NBC)}} \\
Noise $\downarrow$           & 0.494 & 0.352 & 0.300 & \textbf{0.282} \\
Transition $\downarrow$      & 0.385 & 0.305 & \textbf{0.169} & 0.186 \\
Entropy $\uparrow$           & 0.651 & 0.714 & 0.696 & \textbf{0.719} \\
Gini $\downarrow$            & \textbf{0.146} & 0.156 & 0.167 & 0.167 \\
\# Topics $\uparrow$         & 2.73 & 2.48 & 2.87 & \textbf{2.96} \\
\bottomrule
\end{tabular}
\caption{Topic structure. \textbf{T+A}/\textbf{T+V} = bi-modal concatenation baselines (no gate, no interaction terms).}
\label{tab:structure}
\end{table}

The bi-modal baselines reveal an instructive pattern, Text+Video alone closes most of the gap from Text-Only to \Witness on structural metrics (e.g., German noise: 0.273 $\rightarrow$ 0.087 vs.\ \Witness's 0.057), confirming that the visual stream is the dominant multimodal signal.
Text+Audio, in contrast, actively degrades several metrics -- Gini rises from 0.384 to 0.421 and the topic count collapses from 19.1 to 10.1 on German -- indicating that naive audio concatenation conflates acoustic with topical similarity.
\Witness's similarity gate adds a small but consistent improvement over T+V by integrating audio safely: the gate's contribution is best understood as preventing audio from dominating the partition rather than as the sole driver of multimodal gains.
On English Transition Rate, T+V (0.169) outperforms \Witness (0.186); we
attribute this to the much shorter broadcasts (75 vs.\ 411 segments per
video), where audio's marginal value over a vision-led partition is smaller
and over-merging is harder to penalize.

\subsection{Embedding-Space Cluster Quality}
\autoref{tab:cluster} reports clustering validity across representation spaces.
Under text-based topic assignment, none of the embedding spaces yields well-separated clusters, as indicated by negative Silhouette values and low CH indices.
In contrast, topic labels derived from multimodal similarity-gated fusion yield stronger cluster separation across all spaces: Silhouette scores become positive except in the audio space, where they move substantially toward zero, CH indices increase by factors of 4--12 on German and 3--8 on English, and DB indices decrease substantially.
\textcolor{black}{
The bi-modal split reveals where the geometric structure lives. In the visual space, T+V alone achieves the tightest clusters (CH 33.37, Sil 0.42 on German; CH 24.63, Sil 0.40 on English), even slightly above \Witness. 
\Witness, however, dominates the fused space (CH 23.98 vs.\ T+V's 18.82 on German; CH 15.33 vs.\ 13.05 on English), indicating that the \MMTM trades a small amount of visual-space tightness for substantially better joint-space integration.
This is consistent with the structural metrics: visual evidence drives most of the partitioning, and the gate's role is to align audio with the visual-text joint geometry rather than to add separate visual structure.}
%

\begin{table}[t]
\centering
\footnotesize
\setlength{\tabcolsep}{3pt}
\begin{tabular}{l rrrr}
\toprule
\textbf{Metric} & \textbf{Text} & \textbf{T+A} & \textbf{T+V} & \textbf{\Witness} \\
\midrule
\multicolumn{5}{l}{\textit{German}} \\
CH (Visual) $\uparrow$       & 2.81  & 12.52 & \textbf{33.37} & 31.89 \\
Sil. (Visual) $\uparrow$     & -0.13 & -0.01 & \textbf{0.42}  & 0.36 \\
CH (Fused) $\uparrow$        & 2.70  & 23.24 & 18.82 & \textbf{23.98} \\
Sil. (Fused) $\uparrow$      & -0.14 & 0.12  & 0.26  & \textbf{0.33} \\
\midrule
\multicolumn{5}{l}{\textit{English (NBC)}} \\
CH (Visual) $\uparrow$       & 3.95  & 9.06  & \textbf{24.63} & 21.93 \\
Sil. (Visual) $\uparrow$     & 0.02  & 0.11  & \textbf{0.40}  & 0.37 \\
CH (Fused) $\uparrow$        & 3.49  & 11.78 & 13.05 & \textbf{15.33} \\
Sil. (Fused) $\uparrow$      & 0.04  & 0.22  & 0.28  & \textbf{0.34} \\
\bottomrule
\end{tabular}
\caption{Cluster validity in visual and fused embedding spaces. CH = Calinski-Harabasz (higher = tighter, more separated); Sil. = Silhouette ($[-1, 1]$, higher better). T+V leads in visual space; \Witness leads in the joint space.}
\label{tab:cluster}
\end{table}

\subsection{Coherence and Semantic Alignment}
Beyond geometric improvements, lexical coherence (NPMI) increases from 0.77 to 0.86, while topic diversity remains very high (0.998 vs. 0.999).
This suggests that the multimodal model improves both lexical interpretability and structural properties.
Image embedding coherence (IEC) \cite{gonzalez2024neural} increases substantially in the fused space, from 0.59 to 0.79, suggesting that the topics are better aligned with visual semantics when guided by multimodal evidence.
A similar pattern is observed in the audio space, where the IEC score increases from 0.70 to 0.80.
While word embedding (WE) alignment \cite{gonzalez2024neural} remains stable at around 0.30, the overall semantic structure becomes tighter.
\textcolor{black}{The bi-modal baselines reveal that each modality maximizes alignment in its own dominant space: T+A leads in audio-space IEC on English (0.820), while T+V leads in visual-aligned metrics on both corpora.}
\textcolor{black}{Only \Witness wins or ties on fused-space IEC on both German (0.793) and English (0.797), indicating that tri-modal fusion is what produces joint cross-modal alignment rather than modality-specific alignment.}
\textcolor{black}{On English, NPMI is essentially flat (0.631 vs.\ 0.612), while topic diversity rises substantially (0.723 to 0.802); we attribute the flat NPMI to the much shorter average broadcast length on NBC ($\approx$5.4 vs.\ $\approx$25 min for \textit{Tagesschau}), which yields fewer segments per video and a smaller topic vocabulary, limiting headroom for lexical coherence gains.}

\begin{table}[t]
\centering
\footnotesize
\setlength{\tabcolsep}{3pt}
\begin{tabular}{l rrrr}
\toprule
\textbf{Metric} & \textbf{Text} & \textbf{T+A} & \textbf{T+V} & \textbf{\Witness} \\
\midrule
\multicolumn{5}{l}{\textit{German}} \\
NPMI $\uparrow$              & 0.768 & 0.678 & 0.853 & \textbf{0.861} \\
Diversity $\uparrow$         & 0.998 & 0.988 & \textbf{0.999} & \textbf{0.999} \\
WE (Fused) $\uparrow$        & \textbf{0.305} & 0.277 & 0.286 & 0.286 \\
IEC (Fused) $\uparrow$       & 0.591 & 0.675 & 0.740 & \textbf{0.793} \\
IEC (Audio) $\uparrow$       & 0.700 & 0.777 & 0.734 & \textbf{0.800} \\
\midrule
\multicolumn{5}{l}{\textit{English (NBC)}} \\
NPMI $\uparrow$              & \textbf{0.631} & 0.591 & 0.606 & 0.612 \\
Diversity $\uparrow$         & 0.723 & 0.790 & 0.780 & \textbf{0.802} \\
WE (Fused) $\uparrow$        & \textbf{0.237} & 0.224 & 0.222 & 0.221 \\
IEC (Fused) $\uparrow$       & 0.736 & 0.776 & 0.793 & \textbf{0.797} \\
IEC (Audio) $\uparrow$       & 0.766 & \textbf{0.820} & 0.781 & 0.799 \\
\bottomrule
\end{tabular}
\caption{Semantic quality and cross-modal alignment. NPMI/Diversity = lexical interpretability; WE/IEC = alignment of topic words with word- and image-embedding spaces.}
\label{tab:semantic}
\end{table}

\subsection{Cross-Lingual Generalization}
\label{sec:english_results}
Across all three families of metrics (\autoref{tab:structure},\autoref{tab:cluster}, \autoref{tab:semantic}), the structural and geometric gains transfer cleanly from German to English: a 43\% noise reduction, a halved transition rate, and a 5.5$\times$ improvement in visual-space CH. Lexical coherence gains do not transfer, consistent with the corpus-density explanation above. 
We conclude that the improvements achieved by \Witness reflect a general property of similarity-gated tri-modal fusion -- robust to language, broadcaster, and editorial format -- while NPMI specifically depends on per-video segment volume.

\section{Ablation Study}
\label{sec:ablation}
 
\begin{figure}[h]
\centering
\includegraphics[width=\linewidth]{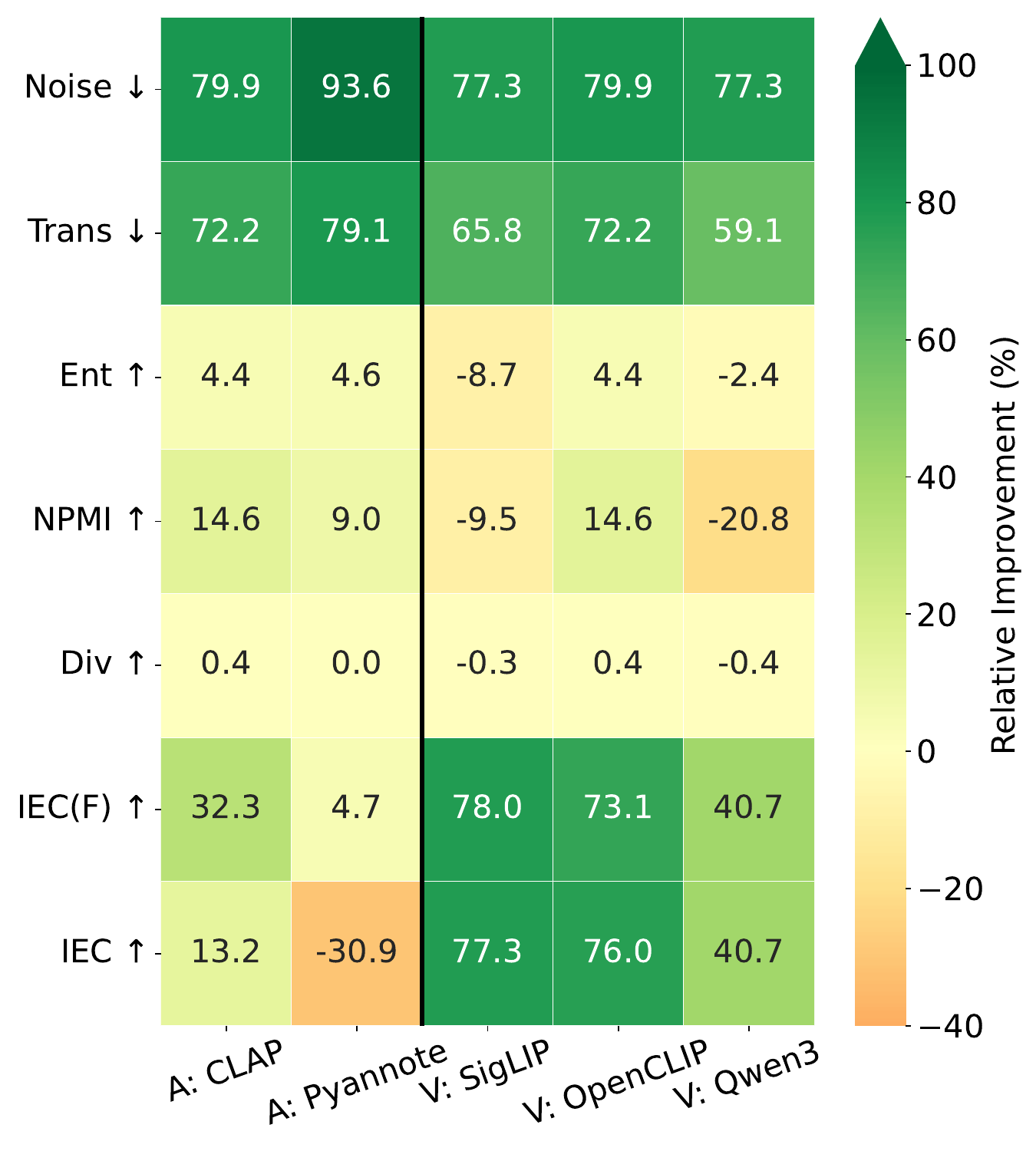}
\caption{Relative improvement (\%) of multimodal encoders over text-only baseline. $\downarrow$/$\uparrow$: lower/higher better.}
\label{fig:combined_ablation}
\end{figure}
 
We ablate encoders on a stratified 15-day subsample ($\approx$9 hours, $\approx$17\% of the German corpus); baseline results on this subset match the full corpus, indicating representativeness. Detailed numerical results are in ~\autoref{sec:appendix_ablation}.
 
\paragraph{Visual encoders.} Comparing OpenCLIP, SigLIP \cite{zhai:et:al:2023:siglip}, and Qwen3-VL-Embedding-8B \cite{li:et:al:2026qwen3vl} (vLLM-served) with all else held constant: all three outperform the text-only baseline on structural metrics, so visual integration is robust to encoder choice. OpenCLIP achieves the best balance (noise 0.063, transition 0.201, NPMI 0.866). SigLIP yields the highest IEC Fused (0.808). Qwen3-VL, despite being substantially larger, underperforms on NPMI (0.599) and IEC (0.639), suggesting that richer visual-semantic representations do not straightforwardly improve topic coherence under parameter-free fusion.
 
\paragraph{Audio backends.} Comparing CLAP and Pyannote speaker embeddings \cite{Bredin23,Plaquet23} with OpenCLIP fixed: both improve over text-only. Pyannote achieves the lowest noise (2.0\%) and transition rate (0.151)---speaker-coherent segmentation---but lower NPMI (0.824 vs.\ 0.866) and much weaker audio-space alignment (IEC 0.488 vs.\ 0.799). This is expected, as Pyannote encodes speaker identity rather than acoustic scene semantics.
 
\section{Human Annotation Validation}
We validate topic interpretability through a two-stage protocol following \citet{Lokmanoglu:et:al:2025:validation}, with five annotators across 71 data points.
 
\paragraph{Stage 1: LLM-driven labeling.} We use \texttt{gemma3:latest} \cite{gemma3:team:2025} via Ollama \cite{ollamaGemma3} to generate cluster labels from top-$k$ keywords and representative excerpts as semantic anchors. This step does not influence topic discovery.
 
\paragraph{Stage 2: Dual-annotator visual evaluation.} We adapt image-intrusion and topic-matching tests to the visual domain.
 
\textit{Image intrusion:} annotators identify an intruder among topic-representative images plus a distractor. Across 114 judgments, overall accuracy is 71.9\%. Inter-annotator agreement on 43 doubly-annotated pairs yields $\kappa = 0.12$ with raw pair agreement of 62.8\%; this low $\kappa$ reflects task design (annotators may identify the same intruder but choose different distractors) rather than poor topic quality. Confusion concentrates on visually homogeneous studio segments.
 
\textit{Topic matching:} coders match a target image to one of four topic rows and rate coherence on a 1--5 scale. Across 57 doubly-annotated pairs, mean coherence is 3.04 ($\text{SD}{=}1.35$), with weighted $\kappa = 0.42$ (moderate), ICC$(2,1){=}0.61$, and Spearman $r_s{=}0.61$ ($p<0.001$)---acceptable reliability \citep{shrout1979intraclass}. Near-agreement ($\leq$1 point) holds in 73.7\% of pairs. Coherence is high for visually distinctive topics (conflict, weather, sports: up to 5.0) and low for studio talking-head scenes (near 1.0), consistent with automated IEC patterns.
 
\section{Discussion}

Our findings reveal two distinct facets of topic quality: \textit{structural coherence} (noise ratio, transition rate, entropy, cluster validity) and \textit{lexical interpretability} (NPMI, diversity). On the German corpus, multimodal fusion improves both: noise drops from 0.27 to 0.06, transition rate from 0.70 to 0.21, and NPMI rises from 0.77 to 0.86. On NBC News, structural and geometric gains transfer cleanly---43\% lower noise, halved transition rate, $5.5\times$ visual-space CH---while lexical coherence remains flat (0.631 vs.\ 0.612). This divergence suggests structural improvement is the robust effect of fusion, while lexical gains depend on per-video segment volume.

\paragraph{Where cross-modal structure resides.} Under text-only assignment, all embedding spaces yield negative Silhouette scores and low CH indices. Under multimodal assignment, all spaces become structured, with the strongest gains in the visual space (CH 2.81$\rightarrow$31.89 on German; 3.95$\rightarrow$24.63 on English), suggesting fusion induces a geometrically coherent joint partitioning rather than merely projecting text topics onto visual evidence. T+V alone achieves the tightest visual-space clusters, while \Witness dominates the fused space, confirming that the gate aligns audio with the visual-text joint geometry.

\paragraph{Encoder trade-offs.} OpenCLIP best balances structural and lexical metrics; SigLIP yields the strongest cross-modal alignment (IEC Fused 0.808); Qwen3-VL, despite its larger scale, underperforms on NPMI and IEC, indicating that richer visual-semantic representations do not automatically improve topic structure under parameter-free fusion. The CLAP--Pyannote comparison shows a parallel trade-off between structural tidiness and semantic alignment.

\paragraph{Limits of the similarity gate.} The fusion is intentionally parameter-free. This suffices with well-pretrained encoders, but cannot adaptively route information by segment content. Studio segments, where the visual stream is uninformative, would benefit from down-weighting visual evidence. Adaptive weighting, content-aware gating, and uncertainty-aware fusion are natural next steps.

\paragraph{Scope.} \Witness is evaluated in a seed-based inference-only setting on news corpora spanning two languages, broadcasters, and editorial formats. As no prior method performs joint tri-modal topic modeling over continuous video, text-only BERTopic under identical conditions is the strongest comparable baseline.
Generalization to multi-genre content remains untested; the open-source release is intended to support community evaluation.

\section{Conclusion}

We introduced \Witness, a modular pipeline for topic discovery in long-form video that integrates speech transcription, audio embeddings, and visual representations through similarity-gated fusion and BERTopic clustering. Evaluated cross-lingually on German (Tagesschau, ${\sim}$54\,h) and English (NBC News, ${\sim}$20\,h), tri-modal fusion substantially improves structural topic quality: noise drops from 0.27 to 0.06, transition rate from 0.70 to 0.21, and entropy rises from 0.84 to 0.92. Cluster validity improves by 5--12$\times$ across embedding spaces. Lexical coherence (NPMI) rises from 0.77 to 0.86 on longer German broadcasts but is corpus-dependent and does not transfer to the shorter NBC format.

Ablation shows visual integration is the dominant source of gain, while the gate prevents audio from dominating the partition. Human annotation provides convergent validation: annotator coherence judgments align with automated IEC scores, with visually distinctive topics rated highest by both. The parameter-free gate's inability to adaptively weight modalities by content remains the main limitation; generalization beyond broadcast news is untested. We release the pipeline code, the 54-hour multimodal corpus with dual-annotator evaluation, and the annotation toolkit to support reproducible multimodal topic modeling research.

\section*{Limitations}
The pipeline inherits standard ASR vulnerabilities, including substitution and deletion errors in noisy or accented speech, and may over-segment semantically continuous discourse at weak lexical boundaries.
The similarity-gated fusion uses fixed modality weights, which may be suboptimal for domains where one modality is systematically less informative; dynamic or learned weighting remains an important future direction.
Guided seeding increases controllability but can bias topic discovery toward predefined concepts, potentially suppressing emergent themes not anticipated by the seed vocabulary.
\textcolor{black}{The evaluation covers two languages (German, English) and two broadcasters across broadcast news, but generalization to other genres (e.g., lectures, interviews, user-generated video) or less structured video formats has not yet been established.}
\textcolor{black}{The lexical-coherence gains we report are corpus-dependent: NPMI improves substantially on the longer-format German corpus but is essentially flat on the shorter NBC broadcasts, suggesting that per-video segment volume is a relevant factor we do not yet control for.}
As no existing method performs joint tri-modal topic modeling over continuous video, direct method-to-method comparison is not possible; we consider establishing such a benchmark a contribution in itself.

\section*{Ethical Considerations}
\paragraph{Data and privacy.}
\textcolor{black}{Videos were collected from the official \texttt{Tagesschau} public broadcast archive and from the publicly released NBC News corpus of \citet{lokmanoglu2025vistopics}, both for non-commercial research purposes.}
Audio and visual features are stored as dense embeddings that cannot be used to reconstruct the original material.

\paragraph{Human annotation.}
Annotators evaluated only short visual stimuli with no sensitive personal information. No demographic or identifying data were collected from participants.
\paragraph{LLM-assisted labeling.}
Topic labels generated by \texttt{gemma3:latest} serve solely as interpretability aids for human annotators and do not influence topic discovery. As with any LLM-generated output, labels may reflect biases in the model's training data and should be treated as suggestions rather than authoritative descriptions.
\paragraph{Encoder bias.}
The pretrained encoders (Whisper, CLAP, OpenCLIP) were trained on large-scale web corpora and may carry societal biases. We encourage future work to audit topic assignments for systematic disparities across speaker demographics and visual content.
\paragraph{Potential misuse.}
We release code and data (aligned with \texttt{Tagesschau} policy) for research use only. Users should assess relevant legal and ethical constraints before deploying the pipeline in applied settings involving
political or sensitive broadcast content.

\section*{Acknowledgments}
This research is partially funded by \Anon

\section*{Disclaimer}
 During the paper writing process, we employed a large language model solely to polish individual sentences for improved readability; no original ideas, data, or analyses were generated by the AI. All intellectual contributions remain the authors’ own work.

\bibliography{custom}

\begin{thebibliography}{35}
\providecommand{\natexlab}[1]{#1}

\bibitem[{Blei and Jordan(2003)}]{blei2003modeling}
David~M Blei and Michael~I Jordan. 2003.
\newblock Modeling annotated data.
\newblock In \emph{Proceedings of the 26th annual international ACM SIGIR conference on Research and development in informaion retrieval}, pages 127--134.

\bibitem[{Blei et~al.(2003)Blei, Ng, and Jordan}]{blei2003latent}
David~M Blei, Andrew~Y Ng, and Michael~I Jordan. 2003.
\newblock Latent dirichlet allocation.
\newblock \emph{Journal of machine Learning research}, 3(Jan):993--1022.

\bibitem[{Bredin(2023)}]{Bredin23}
Hervé Bredin. 2023.
\newblock {pyannote.audio 2.1 speaker diarization pipeline: principle, benchmark, and recipe}.
\newblock In \emph{Proc. INTERSPEECH 2023}.

\bibitem[{Budzinski et~al.(2021)Budzinski, Gaenssle, and Lindst{\"a}dt-Dreusicke}]{Budzinski2021-sl}
Oliver Budzinski, Sophia Gaenssle, and Nadine Lindst{\"a}dt-Dreusicke. 2021.
\newblock The battle of {YouTube}, {TV} and netflix: an empirical analysis of competition in audiovisual media markets.
\newblock \emph{SN Business \& Economics}, 1(9):116.

\bibitem[{Carbonell and Goldstein(1998)}]{Carbonell:Goldstein:1998}
Jaime Carbonell and Jade Goldstein. 1998.
\newblock \href {https://doi.org/10.1145/290941.291025} {The use of mmr, diversity-based reranking for reordering documents and producing summaries}.
\newblock In \emph{Proceedings of the 21st Annual International ACM SIGIR Conference on Research and Development in Information Retrieval}, SIGIR '98, page 335–336, New York, NY, USA. Association for Computing Machinery.

\bibitem[{Cherti et~al.(2023)Cherti, Beaumont, Wightman, Wortsman, Ilharco, Gordon, Schuhmann, Schmidt, and Jitsev}]{cherti2023reproducible}
Mehdi Cherti, Romain Beaumont, Ross Wightman, Mitchell Wortsman, Gabriel Ilharco, Cade Gordon, Christoph Schuhmann, Ludwig Schmidt, and Jenia Jitsev. 2023.
\newblock Reproducible scaling laws for contrastive language-image learning.
\newblock In \emph{Proceedings of the IEEE/CVF Conference on Computer Vision and Pattern Recognition}, pages 2818--2829.

\bibitem[{Elizalde et~al.(2022)Elizalde, Deshmukh, Ismail, and Wang}]{elizalde:2022:clap}
Benjamin Elizalde, Soham Deshmukh, Mahmoud~Al Ismail, and Huaming Wang. 2022.
\newblock \href {https://arxiv.org/abs/2206.04769} {Clap: Learning audio concepts from natural language supervision}.
\newblock \emph{Preprint}, arXiv:2206.04769.

\bibitem[{Fu et~al.(2024)Fu, Wang, Xin, Zhou, Chen, Ge, Janies, and Zhang}]{fu:et:al:2025:misinformation}
Zhe Fu, Kanlun Wang, Wangjiaxuan Xin, Lina Zhou, Shi Chen, Yaorong Ge, Daniel Janies, and Dongsong Zhang. 2024.
\newblock \href {https://arxiv.org/abs/2409.00022} {Detecting misinformation in multimedia content through cross-modal entity consistency: A dual learning approach}.
\newblock \emph{Preprint}, arXiv:2409.00022.

\bibitem[{GemmaTeam et~al.(2025)GemmaTeam, Kamath, Ferret, Pathak, Vieillard, Merhej, Perrin, Matejovicova, Ramé, Rivière, Rouillard, Mesnard, Cideron, bastien Grill, Ramos, Yvinec, Casbon, Pot, Penchev, Liu, Visin, Kenealy, Beyer, Zhai, Tsitsulin, Busa-Fekete, Feng, Sachdeva, Coleman, Gao, Mustafa, Barr, Parisotto, Tian, Eyal, Cherry, Peter, Sinopalnikov, Bhupatiraju, Agarwal, Kazemi, Malkin, Kumar, Vilar, Brusilovsky, Luo, Steiner, Friesen, Sharma, Sharma, Gilady, Goedeckemeyer, Saade, Feng, Kolesnikov, Bendebury, Abdagic, Vadi, György, Pinto, Das, Bapna, Miech, Yang, Paterson, Shenoy, Chakrabarti, Piot, Wu, Shahriari, Petrini, Chen, Lan, Choquette-Choo, Carey, Brick, Deutsch, Eisenbud, Cattle, Cheng, Paparas, Sreepathihalli, Reid, Tran, Zelle, Noland, Huizenga, Kharitonov, Liu, Amirkhanyan, Cameron, Hashemi, Klimczak-Plucińska, Singh, Mehta, Lehri, Hazimeh, Ballantyne, Szpektor, Nardini, Pouget-Abadie, Chan, Stanton, Wieting, Lai, Orbay, Fernandez, Newlan, yeong Ji, Singh, Black, Yu, Hui, Vodrahalli,
  Greff, Qiu, Valentine, Coelho, Ritter, Hoffman, Watson, Chaturvedi, Moynihan, Ma, Babar, Noy, Byrd, Roy, Momchev, Chauhan, Sachdeva, Bunyan, Botarda, Caron, Rubenstein, Culliton, Schmid, Sessa, Xu, Stanczyk, Tafti, Shivanna, Wu, Pan, Rokni, Willoughby, Vallu, Mullins, Jerome, Smoot, Girgin, Iqbal, Reddy, Sheth, Põder, Bhatnagar, Panyam, Eiger, Zhang, Liu, Yacovone, Liechty, Kalra, Evci, Misra, Roseberry, Feinberg, Kolesnikov, Han, Kwon, Chen, Chow, Zhu, Wei, Egyed, Cotruta, Giang, Kirk, Rao, Black, Babar, Lo, Moreira, Martins, Sanseviero, Gonzalez, Gleicher, Warkentin, Mirrokni, Senter, Collins, Barral, Ghahramani, Hadsell, Matias, Sculley, Petrov, Fiedel, Shazeer, Vinyals, Dean, Hassabis, Kavukcuoglu, Farabet, Buchatskaya, Alayrac, Anil, Dmitry, Lepikhin, Borgeaud, Bachem, Joulin, Andreev, Hardin, Dadashi, and Hussenot}]{gemma3:team:2025}
GemmaTeam, Aishwarya Kamath, Johan Ferret, Shreya Pathak, Nino Vieillard, Ramona Merhej, Sarah Perrin, Tatiana Matejovicova, Alexandre Ramé, Morgane Rivière, Louis Rouillard, Thomas Mesnard, Geoffrey Cideron, Jean bastien Grill, Sabela Ramos, Edouard Yvinec, Michelle Casbon, Etienne Pot, Ivo Penchev, and 197 others. 2025.
\newblock \href {https://arxiv.org/abs/2503.19786} {Gemma 3 technical report}.
\newblock \emph{Preprint}, arXiv:2503.19786.

\bibitem[{Gonz{\'a}lez-Pizarro and Carenini(2024)}]{gonzalez2024neural}
Felipe Gonz{\'a}lez-Pizarro and Giuseppe Carenini. 2024.
\newblock Neural multimodal topic modeling: A comprehensive evaluation.
\newblock In \emph{Proceedings of the 2024 Joint International Conference on Computational Linguistics, Language Resources and Evaluation (LREC-COLING 2024)}, pages 12159--12172.

\bibitem[{Grootendorst(2022)}]{grootendorst2022bertopic}
Maarten Grootendorst. 2022.
\newblock Bertopic: Neural topic modeling with a class-based tf-idf procedure.
\newblock \emph{arXiv preprint arXiv:2203.05794}.

\bibitem[{Hu et~al.(2014)Hu, Liu, Jiang, and Yang}]{hu2014latent}
Pengfei Hu, Wenju Liu, Wei Jiang, and Zhanlei Yang. 2014.
\newblock Latent topic model for audio retrieval.
\newblock \emph{Pattern Recognition}, 47(3):1138--1143.

\bibitem[{Ilharco et~al.(2021)Ilharco, Wortsman, Wightman, Gordon, Carlini, Taori, Dave, Shankar, Namkoong, Miller, Hajishirzi, Farhadi, and Schmidt}]{ilharco_gabriel_2021_5143773}
Gabriel Ilharco, Mitchell Wortsman, Ross Wightman, Cade Gordon, Nicholas Carlini, Rohan Taori, Achal Dave, Vaishaal Shankar, Hongseok Namkoong, John Miller, Hannaneh Hajishirzi, Ali Farhadi, and Ludwig Schmidt. 2021.
\newblock \href {https://doi.org/10.5281/zenodo.5143773} {Openclip}.
\newblock If you use this software, please cite it as below.

\bibitem[{Jia et~al.(2011)Jia, Salzmann, and Darrell}]{jia2011learning}
Yangqing Jia, Mathieu Salzmann, and Trevor Darrell. 2011.
\newblock Learning cross-modality similarity for multinomial data.
\newblock In \emph{2011 international conference on computer vision}, pages 2407--2414. IEEE.

\bibitem[{Kim et~al.(2009)Kim, Narayanan, and Sundaram}]{kim2009acoustic}
Samuel Kim, Shrikanth Narayanan, and Shiva Sundaram. 2009.
\newblock Acoustic topic model for audio information retrieval.
\newblock In \emph{2009 IEEE Workshop on Applications of Signal Processing to Audio and Acoustics}, pages 37--40. IEEE.

\bibitem[{Li et~al.(2026)Li, Zhang, Long, Chen, Song, Bai, Yang, Xie, Yang, Liu, Zhou, and Lin}]{li:et:al:2026qwen3vl}
Mingxin Li, Yanzhao Zhang, Dingkun Long, Keqin Chen, Sibo Song, Shuai Bai, Zhibo Yang, Pengjun Xie, An~Yang, Dayiheng Liu, Jingren Zhou, and Junyang Lin. 2026.
\newblock \href {https://arxiv.org/abs/2601.04720} {Qwen3-vl-embedding and qwen3-vl-reranker: A unified framework for state-of-the-art multimodal retrieval and ranking}.
\newblock \emph{Preprint}, arXiv:2601.04720.

\bibitem[{Lokmanoglu and Walter(2025{\natexlab{a}})}]{Lokmanoglu:et:al:2025:validation}
Ayse~D. Lokmanoglu and Dror Walter. 2025{\natexlab{a}}.
\newblock \href {https://doi.org/10.1080/19312458.2025.2549707} {Topic modeling of video and image data: a visual semantic unsupervised approach}.
\newblock \emph{Communication Methods and Measures}, 19(3):232–279.

\bibitem[{Lokmanoglu and Walter(2025{\natexlab{b}})}]{lokmanoglu2025vistopics}
Ayse~D Lokmanoglu and Dror Walter. 2025{\natexlab{b}}.
\newblock Vistopics: A visual semantic unsupervised approach to topic modeling of video and image data.
\newblock \emph{arXiv preprint arXiv:2505.14868}.

\bibitem[{Malzer and Baum(2020)}]{Malzer:2020:hdbscan}
Claudia Malzer and Marcus Baum. 2020.
\newblock \href {https://doi.org/10.1109/mfi49285.2020.9235263} {A hybrid approach to hierarchical density-based cluster selection}.
\newblock In \emph{2020 IEEE International Conference on Multisensor Fusion and Integration for Intelligent Systems (MFI)}, page 223–228. IEEE.

\bibitem[{McInnes et~al.(2020)McInnes, Healy, and Melville}]{mcinnes:et:al:2020:umap}
Leland McInnes, John Healy, and James Melville. 2020.
\newblock \href {https://arxiv.org/abs/1802.03426} {Umap: Uniform manifold approximation and projection for dimension reduction}.
\newblock \emph{Preprint}, arXiv:1802.03426.

\bibitem[{Miao et~al.(2016)Miao, Yu, and Blunsom}]{miao2016neural}
Yishu Miao, Lei Yu, and Phil Blunsom. 2016.
\newblock Neural variational inference for text processing.
\newblock In \emph{International conference on machine learning}, pages 1727--1736. PMLR.

\bibitem[{OllamaTeam()}]{ollamaGemma3}
OllamaTeam.
\newblock gemma3 --- ollama.com.
\newblock \url{https://ollama.com/library/gemma3}.
\newblock [Accessed 12-05-2026].

\bibitem[{Park and Li(2007)}]{park:et:li:2007}
Youngja Park and Ying Li. 2007.
\newblock \href {https://doi.org/10.1109/ICSC.2007.31} {Semantic analysis for topical segmentation of videos}.
\newblock In \emph{International Conference on Semantic Computing (ICSC 2007)}, pages 161--168.

\bibitem[{Plaquet and Bredin(2023)}]{Plaquet23}
Alexis Plaquet and Hervé Bredin. 2023.
\newblock {Powerset multi-class cross entropy loss for neural speaker diarization}.
\newblock In \emph{Proc. INTERSPEECH 2023}.

\bibitem[{Prakash et~al.(2023)Prakash, Wang, Hoang, Hee, and Lee}]{prakash2023promptmtopic}
Nirmalendu Prakash, Han Wang, Nguyen~Khoi Hoang, Ming~Shan Hee, and Roy Ka-Wei Lee. 2023.
\newblock Promptmtopic: Unsupervised multimodal topic modeling of memes using large language models.
\newblock In \emph{Proceedings of the 31st ACM International Conference on Multimedia}, pages 621--631.

\bibitem[{Putthividhy et~al.(2010)Putthividhy, Attias, and Nagarajan}]{putthividhy2010topic}
Duangmanee Putthividhy, Hagai~T Attias, and Srikantan~S Nagarajan. 2010.
\newblock Topic regression multi-modal latent dirichlet allocation for image annotation.
\newblock In \emph{2010 IEEE Computer Society Conference on Computer Vision and Pattern Recognition}, pages 3408--3415. IEEE.

\bibitem[{Qiang et~al.(2017)Qiang, Chen, Wang, and Wu}]{qiang2017topic}
Jipeng Qiang, Ping Chen, Tong Wang, and Xindong Wu. 2017.
\newblock Topic modeling over short texts by incorporating word embeddings.
\newblock In \emph{Pacific-Asia Conference on Knowledge Discovery and Data Mining}, pages 363--374. Springer.

\bibitem[{Radford et~al.(2021)Radford, Kim, Hallacy, Ramesh, Goh, Agarwal, Sastry, Askell, Mishkin, Clark, Krueger, and Sutskever}]{Radford2021LearningTV}
Alec Radford, Jong~Wook Kim, Chris Hallacy, A.~Ramesh, Gabriel Goh, Sandhini Agarwal, Girish Sastry, Amanda Askell, Pamela Mishkin, Jack Clark, Gretchen Krueger, and Ilya Sutskever. 2021.
\newblock Learning transferable visual models from natural language supervision.
\newblock In \emph{ICML}.

\bibitem[{{Radford} et~al.(2022){Radford}, {Kim}, {Xu}, {Brockman}, {McLeavey}, and {Sutskever}}]{radford:2022:whisper}
Alec {Radford}, Jong~Wook {Kim}, Tao {Xu}, Greg {Brockman}, Christine {McLeavey}, and Ilya {Sutskever}. 2022.
\newblock \href {https://doi.org/10.48550/arXiv.2212.04356} {{Robust Speech Recognition via Large-Scale Weak Supervision}}.
\newblock \emph{arXiv e-prints}, arXiv:2212.04356.

\bibitem[{Schuhmann et~al.(2022)Schuhmann, Beaumont, Vencu, Gordon, Wightman, Cherti, Coombes, Katta, Mullis, Wortsman, Schramowski, Kundurthy, Crowson, Schmidt, Kaczmarczyk, and Jitsev}]{schuhmann2022laionb}
Christoph Schuhmann, Romain Beaumont, Richard Vencu, Cade~W Gordon, Ross Wightman, Mehdi Cherti, Theo Coombes, Aarush Katta, Clayton Mullis, Mitchell Wortsman, Patrick Schramowski, Srivatsa~R Kundurthy, Katherine Crowson, Ludwig Schmidt, Robert Kaczmarczyk, and Jenia Jitsev. 2022.
\newblock \href {https://openreview.net/forum?id=M3Y74vmsMcY} {{LAION}-5b: An open large-scale dataset for training next generation image-text models}.
\newblock In \emph{Thirty-sixth Conference on Neural Information Processing Systems Datasets and Benchmarks Track}.

\bibitem[{Shrout and Fleiss(1979)}]{shrout1979intraclass}
P~E Shrout and J~L Fleiss. 1979.
\newblock Intraclass correlations: uses in assessing rater reliability.
\newblock \emph{Psychol Bull}, 86(2):420--428.

\bibitem[{Srivastava and Sutton(2017)}]{srivastava2017autoencoding}
Akash Srivastava and Charles Sutton. 2017.
\newblock Autoencoding variational inference for topic models.
\newblock \emph{arXiv preprint arXiv:1703.01488}.

\bibitem[{Zhai et~al.(2023)Zhai, Mustafa, Kolesnikov, and Beyer}]{zhai:et:al:2023:siglip}
Xiaohua Zhai, Basil Mustafa, Alexander Kolesnikov, and Lucas Beyer. 2023.
\newblock \href {https://arxiv.org/abs/2303.15343} {Sigmoid loss for language image pre-training}.
\newblock \emph{Preprint}, arXiv:2303.15343.

\bibitem[{Zheng et~al.(2014)Zheng, Zhang, and Larochelle}]{zheng2014topic}
Yin Zheng, Yu-Jin Zhang, and Hugo Larochelle. 2014.
\newblock Topic modeling of multimodal data: an autoregressive approach.
\newblock In \emph{Proceedings of the IEEE conference on computer vision and pattern recognition}, pages 1370--1377.

\bibitem[{Zhu et~al.(2013)Zhu, Shyu, and Wang}]{zhu2013videotopic}
Qiusha Zhu, Mei-Ling Shyu, and Haohong Wang. 2013.
\newblock Videotopic: Content-based video recommendation using a topic model.
\newblock In \emph{2013 IEEE International Symposium on Multimedia}, pages 219--222. IEEE.

\end{thebibliography}

\appendix

\section{Appendix: BERTopic Configuration and Seed Topics}
\label{sec:appendix_config}

The main experiments in this paper use the BERTopic settings listed below.
The full pipeline YAML configuration is available in the open‑source repository.
\begin{table}[h]
\centering
\footnotesize
\begin{tabular}{ll}
\toprule
\textbf{Parameter} & \textbf{Value} \\
\midrule
Sentence embedding model & \texttt{all-mpnet-base-v2} \\
Minimum topic size & 5 \\
Merge similarity threshold & 0.70 \\
Embedding source & \texttt{both} (T + MM) \\
Fusion weights (text, audio, visual) & 0.34, 0.33, 0.33 \\
UMAP neighbors / components / metric & 15 / 8 / cosine \\
HDBSCAN min cluster size / metric & 5 / euclidean \\
\bottomrule
\end{tabular}
\caption{BERTopic configuration used in all experiments.}
\label{tab:bertopic_params}
\end{table}

\subsection{Guided Seed Topic Words}
In all guided experiments (\autoref{tab:structure}--\autoref{tab:semantic}), BERTopic is initialized with the following seed topics to steer the discovery toward known semantic regions.

\begin{table}[h]
\centering
\small
\begin{tabular}{ll}
\toprule
\textbf{Theme} & \textbf{Seed words} \\
\midrule
War \& conflict      & war, conflict, battle, army \\
Democracy            & democracy, freedom, election, parliament \\
Peace \& history     & peace, reconciliation, memory, history \\
Economy              & economy, trade, market, growth \\
Climate              & climate, environment, sustainability, green \\
Technology           & technology, innovation, digital, future \\
Health               & health, medicine, pandemic, vaccine \\
Culture              & culture, art, music, literature \\
Sports               & sports, competition, athlete, tournament \\
Education            & education, school, university, learning \\
Human rights         & human rights, justice, equality, activism \\
Migration            & migration, refugee, border, asylum \\
Science              & science, research, discovery, experiment \\
Space                & space, astronomy, exploration, universe \\
Leadership           & leadership, governance, policy, diplomacy \\
\bottomrule
\end{tabular}
\caption{Seed topics used to guide BERTopic in the guided (semi‑supervised) experiments.}
\label{tab:seed_topics}
\end{table}

\section{Encoder Ablation Details}
\label{sec:appendix_ablation}
 
\begin{table}[h]
\centering
\small
\setlength{\tabcolsep}{2pt}
\begin{tabular}{l c|ccc}
\toprule
& \textbf{Text} & \multicolumn{3}{c}{\textbf{Multimodal}} \\
\cmidrule(lr){3-5}
\textbf{Metric} & (shared) & \textbf{SigLIP} & \textbf{OpenCLIP} & \textbf{Qwen3-VL} \\
\midrule
Noise $\downarrow$ & 0.313 & 0.071 & \textbf{0.063} & 0.071 \\
Transition $\downarrow$ & 0.723 & 0.247 & \textbf{0.201} & 0.296 \\
Entropy $\uparrow$ & 0.888 & 0.811 & \textbf{0.927} & 0.867 \\
NPMI $\uparrow$ & 0.756 & 0.684 & \textbf{0.866} & 0.599 \\
Diversity $\uparrow$ & 0.995 & 0.992 & \textbf{0.999} & 0.991 \\
CH (Fused) $\uparrow$ & 2.24 & \textbf{21.0} & 17.6 & 23.4 \\
Sil.\ (Fused) $\uparrow$ & -0.111 & 0.192 & \textbf{0.325} & 0.246 \\
IEC (Fused) $\uparrow$ & 0.454 & \textbf{0.808} & 0.786 & 0.639 \\
\bottomrule
\end{tabular}
\caption{Visual encoder ablation (15-day subsample). Text-only baseline shared.}
\label{tab:ablation_visual}
\end{table}
 
\begin{table}[h]
\centering
\small
\setlength{\tabcolsep}{4pt}
\begin{tabular}{l c|cc}
\toprule
& \textbf{Text} & \multicolumn{2}{c}{\textbf{Multimodal}} \\
\cmidrule(lr){3-4}
\textbf{Metric} & (shared) & \textbf{CLAP} & \textbf{Pyannote} \\
\midrule
Noise $\downarrow$ & 0.313 & 0.063 & \textbf{0.020} \\
Transition $\downarrow$ & 0.723 & 0.201 & \textbf{0.151} \\
NPMI $\uparrow$ & 0.756 & \textbf{0.866} & 0.824 \\
CH (Fused) $\uparrow$ & 2.24 & \textbf{17.6} & 16.3 \\
Sil.\ (Fused) $\uparrow$ & -0.111 & \textbf{0.325} & 0.300 \\
IEC (Audio) $\uparrow$ & 0.706 & \textbf{0.799} & 0.488 \\
\bottomrule
\end{tabular}
\caption{Audio backend ablation (15-day subsample).}
\label{tab:ablation_audio}
\end{table}

\end{document}